\documentclass{llncs}

\begin{document}

\title{SynNet: Structure-Preserving Fully Convolutional Networks for Medical Image Synthesis}
\titlerunning{Hamiltonian Mechanics}  
%
\author{Deepa Gunashekar\inst{1} \and Sailesh Conjeti \inst{1}  \and Abhijit Guha Roy \inst{1} \and Nassir Navab \inst{1} \and Kuangyu Shi \inst{1,2}
}
\authorrunning{Ivar Ekeland et al.} 
%
\tocauthor{Ivar Ekeland, Roger Temam, Jeffrey Dean, David Grove,
Craig Chambers, Kim B. Bruce, and Elisa Bertino}
\institute{ Computer Aided Medical Procedures, Technical University of Munich, Germany\\
\and
Department of Nuclear Medicine, Technical University of Munich, Munich, Germany}

\maketitle              

\begin{abstract}
Cross modal image syntheses is gaining significant interests for its ability to estimate target images of a different modality from a given set of source images,like estimating MR to MR, MR to CT, CT to PET etc, without the need for an actual acquisition.Though they show potential for applications in radiation therapy planning,image super resolution, atlas construction, image segmentation etc.The synthesis results are not as accurate as the actual acquisition.In this paper,we address the problem of multi modal image synthesis by proposing a fully convolutional deep learning architecture called the SynNet.We extend the proposed architecture for various input output configurations. And finally, we propose a structure preserving custom loss function for cross-modal image synthesis.We validate the proposed SynNet and its extended framework on BRATS dataset with comparisons against three state-of-the art methods.And the results of the proposed custom loss function is validated against the traditional loss function used by the state-of-the-art methods for cross modal image synthesis. 

\end{abstract}
\section{Introduction}
The  availability of various imaging modalities such as X-ray, CT, MRI, PET, US etc., for capturing the different tissue characteristics,has led the researchers to look for methods that perform subject specific synthesis, without the need for an actual acquisition.With potential for applications in image super resolution \cite{jog2014improving}\cite{rueda2013single}, building virtual models, atlas construction\cite{roy2014mr}, multimodal registration \cite{degen2016multi},segmentation \cite{wolz2012multi}, and radiation therapy planning \cite{shi2015pet} etc.
The \textit{atlas based method} uses an atlas of source images, for synthesis of target image.By mapping the attenuation maps of a source image to the estimated attenuation map \cite{degen2016multi} \cite{roy2014mr}.The disadvantage of this method is that, the results of the reconstruction is dependent on the registration accuracy. Since the registration quality has a direct impact on the synthesis results.The \textit{data base driven methods} operate by using a database of images with arbitrary source and target modalities for image synthesis. The target modality is estimated, by performing a local patch  based search at each point in the target image using nearest neighbor information value at the given point \cite{ye2013modality}. However, the method restricts nearest neighbor search to a small window and ignores the spatial information, needed for accurate synthesis.The \textit{learning based methods} work by learning a non-linear relationship between the source and the target images for the synthesis task \cite{huynh2016estimating} \cite{vemulapalli2015unsupervised}. Though, these methods outperform the previous methods, the reconstruction quality is dependent on the quality of feature extraction.

Recently, \textit{deep learning models} such as the CNN \cite{nie2016estimating} \cite{ronneberger2015u} \cite{dong2014learning} are gaining a lot of popularity in computer vision and medical imaging fields, for their capability to learn a hierarchy of features.As a multilayer and fully trainable model, CNN’s can capture the non-linear mapping by learning from high level features built upon low level features. The method presented in \cite{nie2016estimating} learns a non-linear mapping from MR to CT images using patch based 3D fully convolutional network. However, since the method is patch based, it does not take into consideration the contextual features of the whole image. Also, the training time and prediction speed of the network is very high.

In this paper, we propose a variant of F-CNN for image synthesis and call it the SynNet. The proposed architecture is end to end without any pre-and post-processing of data and heuristics. The main contributions of this paper is to Investigate an end -to- end fully convolutional network architecture for cross modal image synthesis by using the contextual information of the whole image.Extend the proposed architecture for various input and output configurations.Single input single output (SISO), Multiple input single output (MISO) and Multiple input multiple output (MIMO).
Investigate the proposed custom loss function for cross modal image synthesis.

\section{Problem Statement}
Given an input image \textit{\textbf{I}}, with modality specific information of an underlying anatomy, our task is to synthesize a corresponding target image $\textbf{\textit{S}}^{\textbf{\textit{r}}}$ of the same anatomy, but from a different modality. The target image $\textbf{\textit{S}}^{\textbf{\textit{r}}}$ is reconstructed by learning a data base of 
$\tau$
$\textbf{=} \left \{\textbf{\textit{I}}_{\textbf{\textit{n}}},\textbf{\textit{S}}_{\textbf{\textit{n}}}\right\}_\textbf{\textit{{n=1}}}^\textbf{\textit{{N}}}$ 
containing \textbf{\textit{N}} number of spatially aligned exemplar image pairs of source image $\textbf{\textit{I}}_{\textbf{\textit{n}}}$ and the subject specific image $\textbf{\textit{S}}_{\textbf{\textit{n}}}$.
\subsection{Achitecture}
\label{sec:model}
\begin{figure}
  \centering
  \includegraphics[width=12.0cm]{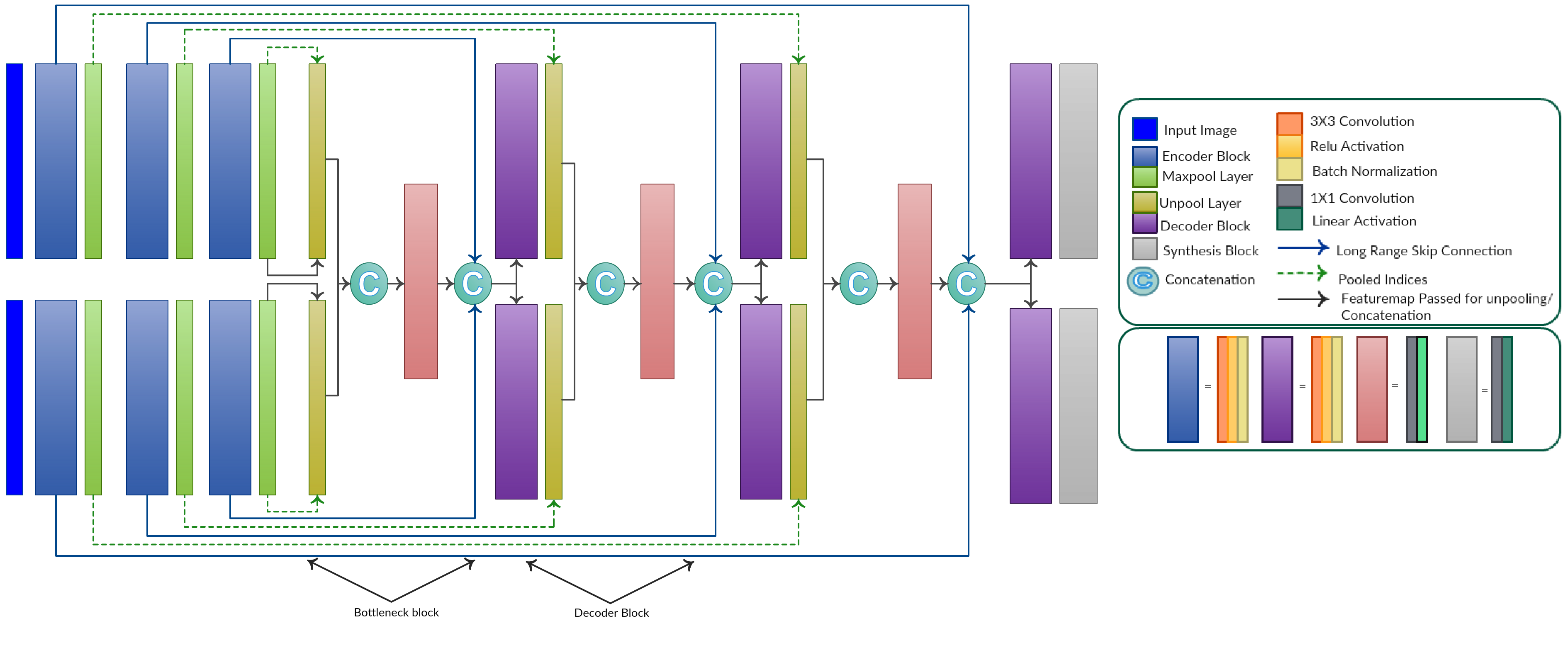}
  \caption{\label{fig:model}Model Architecture:MIMO}
\end{figure}

The proposed SynNet network (Fig.\ref{fig:model}) architecture consists of a contracting path of encoder blocks followed by an expanding path of decoder blocks with long range skip connections for relaying the intermittent feature representations from encoder blocks to their matched decoder blocks through concatenation layers, In particular, we utilize the advantages of the skip connections introduced in U-Net \cite{ronneberger2015u}, and unpooling layers proposed in \cite{noh2015learning}, for upsampling the output of the decoder block for the synthesis task. Further, We extend the SynNet architecture to Single Input Single Output (SISO) consisting of a single encoder and a decoder arm for one to one modality synthesis, Multiple Input Single Output (MISO)  consisting of two encoder arms processed parallely and later concatenated to form a single decoder arm for multi source to single target image synthesis and Multiple Input Multiple Output configurations(MIMO), which consists of two separate arms of encoder blocks followed by two separate arms of decoder block which are later concatenated from the respective matched encoder blocks,for multi target image synthesis.This frame work was designed to evaluate the proposed architecture for synthesis of multi target modality images, by combining the features from multi source modality images.The individual constituent blocks for the proposed architecture are explained in detail in the next section.

\subsubsection{Encoder Block}
\label{sbsec:encoder}
Each encoder block , consists of 4 main layers.A convolution layer consisting of kernels of size $3\times3$ for all the encoder blocks along with bias.A batch normalization layer is introduced after the convolution layer to compensate for the covariate shifts and prevent over fitting during the training procedure.We use the ReLU activation function to introduce  non-linearity in the training. This is followed by a max pooling layer which reduces the height and width of the feature map by half.The pooling indices during max pooling operation are stored which are later used in the unpooling stage of decoder block to preserve spatial consistency\cite{ronneberger2015u}.
\begin{figure}
\begin{minipage}[ht]{1\linewidth}
  \centerline{\includegraphics[width=8.0cm]{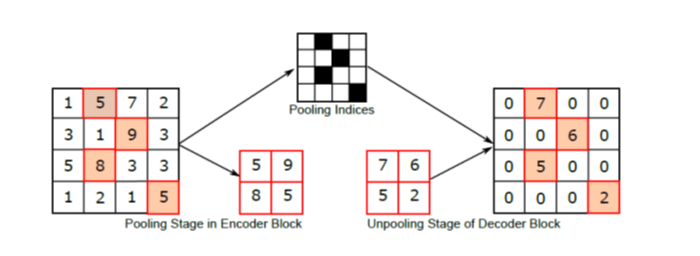}}
  \caption{\label{fig:Unpooling}Illustration of Index based unpooling}
\end{minipage}
\end{figure}
\subsubsection{Decoder Block}
Each decoder block consists of 5 main layers.The unpooling layer,which utilizes the saved pooling indices of the matched encoder block for upsampling the lower resolution feature map from the previous decoder block, to a higher resolution one.This process is illustrated schematically in \ref{fig:Unpooling}.Here we ensure that the spatial information remains preserved, in contrast to using bilinear interpolation for upsampling \cite{noh2015learning}.This is particularly important for the synthesis task, as bilinear interpolation could potentially lead to diffused boundaries,leading to unreliable estimation of edges during reconstruction.The unpooling layer is followed by a concatenation layer,which uses the long skip connections to concatenate the feature maps from the matched encoder block with the unpooled feature maps in the corresponding decoder block. The advantage of such a skip connection is that,it helps in the transition from low to high dimension by adding information rich feature map from encoder, and aides in the flow of gradients to the shallow part of the encoder, thus minimizing the risk of vanishing gradient training deep models \cite{ecb} \cite{drozdzal2016importance}. The concatenation layer is followed by convolutional layer, batch normalization and ReLU activation. The kernel size of convolution layer is kept the same to $3\times3$ to match with the kernel size of the encoder blocks.

\subsubsection{Synthesis Block}
The synthesis block consists of a $1\times1$ convolutional layer to transfer the $64$ dimension feature map to a corresponding sized output dimension, suitable for the synthesis task.This is followed by a linear activation layer, which simply maps the output of the pre-activation to itself.

\section{Cost Function}
In this  section  we discuss the loss function proposed for training the SynNet.

\subsubsection*{$L_{2}$ Loss} The $L_{2}$ loss is the traditional and frequently used loss function for training deep networks for image synthesis. The $L_2$ loss measures the mean squared error between the predictions and ground truth,and is given by the following equation: \begin{equation}
\Im \, _{L_{2}} = \sum_{n=1}^{N}\left \| S_{n} - S_{n}^{r}\right \|_{2} 
\label{eq:reconstruction_loss}
\end{equation} , where $S_{n}$ and $ S_{n}^{r}$ are the ground truth and predicted images, respectively.
The L2 loss penalizes larger errors, but is tolerant to small errors regardless of the underlying structure in the image, while training. Thus it fails to capture the finer characteristics of Human Visual System (HVS) which is sensitive to the luminance, contrast and structure of the image $add ref$.
\subsubsection*{Perceptual Loss: SSIM}
To produce visually pleasing results, the error function must be perceptually motivated. Thus we introduce a perceptually motivated loss,The SSIM measures the error between the predicted image and the ground truth image considering into the changes in the local structure,luminance and masking.The SSIM is given by,
$\Im \, _{SSIM}$ is given by ,\begin{equation}
\Im \, _{SSIM} = \sum_{n=1}^{N} 1-SSIM 
\label{eq:Perceptual_loss}
\end{equation}
where, \begin{equation} SSIM = \frac{2\mu _{S_{n}}\, \mu _{S_{n}^{r}} + C_{1}}{\mu_{S_{n}}^{2}+\mu _{s_{n}^{r}}^{2} +C_{1}} \; \cdot \frac{2\sigma  _{S_{n}}\, \sigma _{S_{n}^{r}} + C_{2}}{\sigma_{S_{n}}^{2}+\sigma _{s_{n}^{r}}^{2} +C_{2}}
\label{eq:Perceptual_loss}
\end{equation} where $S_{n}$ and $ S_{n}^{r}$ are the ground truth and predicted images, $\mu{S_{n}}$,$\mu{S_{n}^{r}}$ is the mean, $\sigma{S_{n}}$,$\sigma{S_{n}^{r}}$ is the standard deviation and $C_{1}$,$C_{2}$ are constants.
\subsubsection*{Weighting Scheme}
 The idea behind such a weighing scheme \cite{dollar2013structured} is that, each pixel $x \in \Omega$ is associated with a weight $\omega(x)$, which dictates the relative importance of the pixel while computing the loss.Thus assigning importance to more deemed regions, while estimating the pixels belonging to the edges as they tend to get diffused often.Secondly, gradients arising from small anatomical structures need to be boosted to compensate for the edge imbalance within the training data, by selectively re-weighting their contributions.
  
 \begin{figure}
\begin{minipage}[h]{1.0\linewidth}
  \centerline{\includegraphics[width=10.0cm]{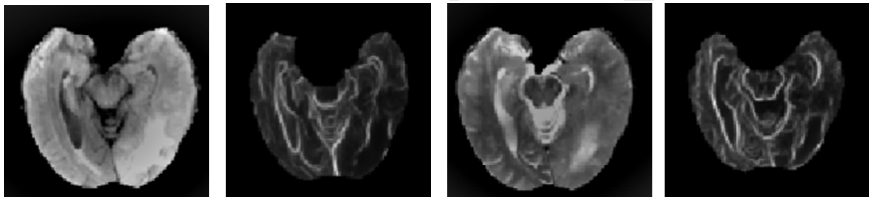}}
  \caption{\label{fig:edgemap}Illustration of Edge weights obtained for T1 and T2 MRI images respectively for training the joint loss function}
\end{minipage}
\end{figure}

Thus, the $L_{2}$ loss in eq (2), and the SSIM loss in eq(3) with the preassigned weight at every pixel location can be rewritten as
\begin{equation}
\Im \, _{L_{2}}^{\omega } = \sum_{n=1}^{N}\sum_{x}\omega\left ( x \right ) \left \| S_{n} - S_{n}^{r}\right \|_{2} 
\label{eq:weighted reconstruction loss}
\end{equation}
 \begin{equation}
\Im \, _{SSIM}^{\omega} = \sum_{n=1}^{N}\sum_{x}\omega \left ( x \right ) \left ( 1-Q\left ( S_{n},S_{n}^{r} \right) \right )
\label{eq:weighted perceptual loss}
\end{equation} where, Q=SSIM respectively.
\subsubsection*{Joint weighted loss}
For the accurate synthesis of the target images,we introduce a novel way of combing the characteristics of the ${L}_{2}$ loss, with the SSIM loss and call it the Joint Loss Function. The proposed loss functions use the weighted combination of the L2 loss function and the SSIM loss function, along with the TV regularization as a smoothening term. The joint loss function for training  of the proposed SynNet, is given by
\begin{equation}
\Im _{overall} = \lambda _{1} \Im\, _{L_{2}}^{\omega } + \lambda _{2} \Im \,_{SSIM}^{\omega } + \lambda _{3} T_{TV} + \lambda _{4} R_{W}
\label{eq:cost_joint}
\end{equation}
Where $\Im\, _{L_{2}}^{\omega }$ is the weighted $L_{2}$ loss from eq(3), $\Im \,_{SSIM}^{\omega }$ is the weighted SSIM loss from eq(4), $T_{TV}$ is the total variation, given by
\begin{equation}
T_{TV} = \sum_{n=1}^{N}\sum_{i,j}t\left ( i,j \right )
\label{eq:total_variation}
\end{equation} 
where n is the number of images in the training set and $\left(i,j\right)$ is the rows and columns at every pixel location, and
\begin{equation}
t\left ( i,j \right )=\sqrt{p^{2}+q^{2}}  
\label{eq:TV_at every pixel location}
\end{equation}
where $p = S_{n}^{r}\left ( i+1,j \right )-S_{n}^{r}\left ( i,j \right )$ and $q = S_{n}^{r}\left ( i,j+1 \right )-S_{n}^{r}\left ( i,j \right)$. The term $R_{w}$ is the relative weight, $\lambda _{1}, \lambda _{2}, \lambda _{3}, \lambda _{4}$ are the relative weight for the individual loss functions.

\subsection{Optimization}
In this section, we discuss the optimization of the proposed cost function for training the SynNet. The training problem is to optimize the joint cost function to estimate the weights and the bias $\Theta =\{W_{\left(\cdot\right)},b_{\left(\cdot\right)}\}$ associated with all the layers.

\begin{equation}
\Theta_{\{W_{\left(\cdot\right)},b_{\left(\cdot\right)}\}}^{\star} = \mathrm{argmin_{\,\Theta}} \, \Im _{Overall}^{\,\omega} \left(\Theta / \tau \right)
\end{equation}
Where, $\Theta^{\star}$ is the optimal parameter set that minimizes the overall cost function. This cost function is optimized using stochastic gradient descent and back propagation. The derivative of the cost function w.r.t the parameters is given by
\begin{equation}
\frac{\partial\Im_{Overall}^{\,\omega}}{\partial\Theta} = \frac{\partial\Im_{Overall}^{\,\omega}}{\partial S_{n}} \, \frac{\partial S_{n}}{\partial \Theta}
\end{equation}
The second term,$\frac{\partial S_{n}}{\partial \Theta}$  is estimated via chain rule by back propagating the gradients. The first term $\frac{\partial\Im_{Overall}^{\,\omega}}{\partial S_{n}}$
first term,  is estimated as:
\begin{equation}
\frac{\partial\Im_{Overall}^{\,\omega}}{\partial S_{n}} =\lambda_{1}\frac{\partial\Im_{L_{2}}^{\,\omega}}{\partial S_{n}}\, + \lambda_{2}\frac{\partial\Im_{SSIM}^{\,\omega}}{\partial S_{n}}\,+\lambda_{3}\frac{\partial T_{TV}}{\partial S_{n}} 
\end{equation}
where the derivatives of the individual cost function is given by
\begin{equation}
\frac{\partial\Im_{L_{2}}^{\,\omega}}{\partial S_{n}} = -2\omega\left(x\right)\left(S_{n}\left(x\right)-S_{n}^{r}\left(x\right)\right)
\end{equation}
For the sake of simplicity, we split the terms in the eq(3) as
\begin{equation}
l_{n} = \frac{2\mu _{S_{n}}\, \mu _{S_{n}^{r}} + C_{1}}{\mu_{S_{n}}^{2}+\mu _{s_{n}^{r}}^{2} +C_{1}}  ,\; c_{n} = \frac{2\sigma  _{S_{n}}\, \sigma _{S_{n}^{r}} + C_{2}}{\sigma_{S_{n}}^{2}+\sigma _{s_{n}^{r}}^{2} +C_{2}}
\end{equation}
The derivative of the SSIM loss is estimated
\begin{equation}
\frac{\partial\Im_{SSIM}^{\,\omega}}{\partial S_{n}} = \frac{\partial l_{n}}{\partial S_{n}} \, + \frac{\partial c_{n}}{\partial S_{n}}
\end{equation}
The derivative of the individual term is calculated using chain rule.
The derivative of the regularization parameter $\frac{\partial \Im_{T_{TV}}}{\partial S_{n}}$ from eq(8), is obtained by
\begin{equation}
\frac{\partial \Im_{T_{TV}}}{\partial S_{n}} = \frac{\partial S_{n\left ( i,\,j \right )}^{r}}{\partial t_{\left ( i,\,j \right )}} + \frac{\partial S_{n\left ( i-1,\,j \right )}^{r}}{\partial t_{\left ( i-1,\,j \right )}} + \frac{\partial S_{n\left ( i,\,j-1 \right )}^{r}}{\partial t_{\left ( i,\,j-1 \right )}}  
\end{equation}
The learning parameters $\Theta$ is updated by mini-batch stochastic gradient descent after every
iteration as
\begin{equation}
\Theta ^{k} = \Theta \,^{k-1} - \eta ^{k}
\label{eq:learning parameter}
\end{equation}
\begin{equation}
\eta ^{k} = \rho \, \eta ^{\, k-1}\, +\gamma \frac{\partial \Im _{overall\:loss}}{\partial \Theta ^{k-1}}
\end{equation}
Here k denotes the current iteration, $\rho$ is the momentum and $\gamma$ is the learning rate for SGD.It should be noted that these two parameters are chosen very carefully to obtain convergence.
%
%
%
\section{Experimental Results and discussions}
\subsection{Dataset}
The proposed SynNet architecture is evaluated on the publicly available BRATS dataset.The dataset consists of 120 low and high resolution T1, T2, contrast enhanced T1 and Flair MR images of size 181x181 obtained from patients suffering from glioma.We divide the data from subject 1-96 as the training set and subject 97-120 as the testing set. Thus, resulting in a total of 96 annotated images for training and 24 annotated images for testing. Further, we augment the datasets by introducing random horizontal flips, vertical flips, rotation, and scaling to create more variability in the training data. The data sets are skull stripped and co-registered to the same anatomical template.

\subsection{Experimental Settings}
All the models were trained on  Intel core-i7 machine with a 12GB Nvidia GeForce GTX 1080 GPU. The code was implemented in a PYTHON based open source library Keras[39] running on top of Theano version (v.0.9).The hyperparameters in loss function in Eq. (13.3) were set as $\lambda_{1}$=10, $\lambda_{2}$ = 5, $\lambda_{3}$ = 0.5 and $\lambda_{4}$ is the weight decay 0.0001 for the training of the SynNet. The SGD optimization is performed in mini batches of size = $32$. The learning rate $\gamma$ is set to 0.01. The training was performed with a momentum of $\rho$ = 0.9. The training parameters are kept constant for all the deep learning comparative methods and baselines defined in next section, for fair comparison. All the networks were run till convergence. The training was conducted with expert 1 annotations and expert 2 annotation is reserved for validation purposes. 
\subsubsection{Comparative methods and baselines}
We define two base lines to discuss the results of the work presented in this paper.The first base lines, BL-1 is used to evaluate the architecture of the proposed SynNet , for multi modal image synthesis against the following state of the art deep learning algorithms. We compare our framework against the three popular state of the art deep learning methods for image synthesis.The comparative methods are listed as follows, CM-1 CNN \cite{nie2016estimating},CM-2 Deconv-net\cite{dong2014learning} architecture, CM-3 U-net \cite{ronneberger2015u} architecture.These comparative methods, defined above uses the L2 loss for image synthesis. The salient aspects of the comparative methods is tabulated in Table~\ref{table:Comparative Methods} .The second baseline, BL-2 Table~\ref{table:BL-2} highlights the performance of the proposed custom loss function in training the SynNet for SISO,MISO and MIMO framework.

\subsubsection{Evaluation Metric}
The performance of the proposed SynNet in comparison to the defined comparative methods and baselines are evaluated based on 2 standard metrics.Peak Signal to Noise Ratio (PSNR) which is the most commonly used performance measure to evaluate the quality of reconstruction. And Structural Similarity Index (SSIM) a metric used in the measurement of image quality based on the structural similarities between the original image and the predicted image.The SSIM measures the image quality in terms of the human observer, by correlating well with the sensitivity of HVS with respect to the luminance, contrast and structure of the predicted image in the visual scene.The higher the SSIM, the better the quality of reconstruction.

\begin{table}[ht]
\caption{Comparative Methods} 
\centering 
\begin{tabular}{c c c c} 
\hline\hline 
CM & Unpool & Upsampling & skip connections\\ [0.5ex] 
\hline 
$CM-1$ & $-$ & $-$ & $-$\\ 
$CM-2$ & $*$ & $-$ & $-$\\
$CM-3$ & $-$ & $*$ & $*$\\
$SynNet$ & $*$ & $-$ & $*$\\[1ex] 
\hline 
\end{tabular}
\label{table:BL-1 Comparative Methods} 
\end{table}
\begin{table}[ht]
\caption{Proposed loss functions for various architectural frameworks} 
\centering 
\begin{tabular}{c c c c} 
\hline\hline 
Loss & SISO & MISO & MIMO \\ [0.5ex] 
\hline 
$L_{2}loss$ & $+$ & $+$ & $+$\\ 
$Weighted L_{2}$ & $*$ & $*$ & $*$\\
$Joint loss$ & $\times$ & $\times$ & $\times$\\[1ex] 
\hline 
\end{tabular}
\label{table:BL-2} 
\end{table}
\subsection{Experimental Results}
We present the results of our work in two parts. First, We present the results of the proposed SynNet along with the comparative methods presented in the Table~\ref{table:Comparative Methods}. Next, we present the results of the SynNet, for the proposed custom loss function for SISO, MISO and MIMO frameworks of the SynNet  and show that the quality of results improves significantly with better loss functions,Table~\ref{table:Proposed loss function} even when the network architecture is unchanged.
\begin{table*}
\begin{center}
\centering
\begin{tabular}{c c c c c c c c c}
\hline\hline
 & Input & Output & PSNR(dB) & SSIM & Input & Output & PSNR(dB) & SSIM \\
\hline
$CM-1$ & $T1$ & $T2$ & $13.89$ & $0.63$ & $T2$ & $T1$ & $15.82$ & $0.74$ \\
$CM-2$ & $T1$ & $T2$ & $19.93$ & $0.54$ & $T2$ & $T1$ & $17.95$ & $0.54$  \\
$CM-3$ & $T1$ & $T2$ & $16.45$ & $0.52$ & $T2$ & $T1$ & $18.93$ & $0.50$ \\ 
$SynNet$ & $T1$ & $T2$ & $20.75$ & $0.65$  & $T1$ & $T1$ & $19.80$ & $0.64$ \\[1ex]
\hline
\end{tabular}
\caption{Results of Comparative methods}\
\label{table:Comparative Methods} 
\end{center}
\end{table*}
\begin{table*}
\begin{center}
\begin{tabular}{c c c c c c c c c }
\hline\hline
Loss & Input & Output & PSNR(dB) & SSIM & Input & Output & PSNR(dB) & SSIM \\
\hline
$L_{2}$ & $T1$ & $T2$  & $20.75$ & $0.65$ & $T2$ & $T1$  & $19.80$ & $0.64$ \\
$Weighted L_{2}$ & $T1$ & $T2$  & $21.80$ & $0.77$ & $T2$ & $T1$  & $20.61$ & $0.69$ \\
$Joint Loss$ & $T1$ & $T2$  & $22.80$ & $0.86$ & $T2$ & $T1$  & $21.65$ & $0.70$\\[1ex]
\hline
$L_{2}$ & $T1,T1c$ & $T2$  & $20.44$ & $0.71$ & $T2,Flair$ & $T1$  & $18.91$ & $0.74$ \\
$Weighted L_{2}$ & $T1,T1c$ & $T2$  & $22.56$ & $0.70$ & $T2,Flair$ & $T1$  & $22.53$ & $0.74$ \\
$Joint Loss$ & $T1,T1c$ & $T2$  & $16.82$ & $0.80$ & $T2,Flair$ & $T1$  & $20.35$ & $0.81$ \\[1ex]
\hline
$L_{2}$ & $T1,T1c$ & $T2,Flair$ & $19.95|17.73$ & $0.70 |
0.60$ & $T2,Flair$ & $T1,T1c$  & $18.33|18.03$ & $0.67|0.62$ \\
$Weighted L_{2}$ & $T1,T1c$ & $T2,Flair$ & $21.05|20.91$ & $0.79|0.72$ & $T2,Flair$ & $T1,T1c$  & $20.07|19.73$ & $0.75|0.73$ \\
$Joint Loss$ & $T1,T1c$ & $T2,Flair$  & $21.05|20.91$ & $0.79|0.81$& $T2,Flair$ & $T1,T1c$  & $20.55|19.78$ & $0.81|0.76$ \\[1ex]
\hline
\end{tabular}
\caption{Results of SynNet with respect to the proposed loss functions}\
\label{table:Proposed loss function}
\end{center}
\end{table*}

\begin{figure}[ht]
  \centerline{\includegraphics[width=10.0cm,height=2cm]{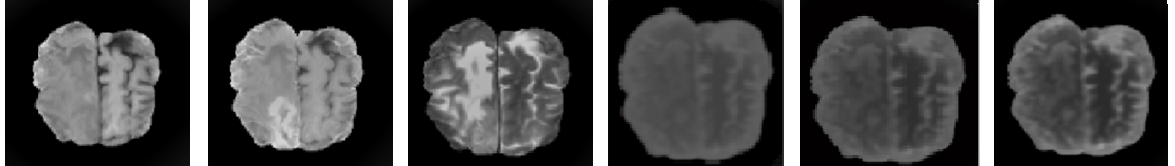}}
  \caption{\label{fig:T2Reconstruction} MISO: Reconstruction of T2 MR for a sample dataset. Inputs T1, T1c MR and Ground truth T2 with $L_{2}$ loss, weighted $L_{2}$ loss and Joint loss functions respectively: We successfully reconstructed the T2 MRI, without the lesions in comparison to the ground truth. Also we can observe that the visual quality of reconstruction obtained using the proposed joint loss function shows prominence in preserving the structure compared to the reconstruction obtained for $L_{2}$ loss and the weighted $L_{2}$loss functions respectively.}
\end{figure}
\subsection{Experimental Observation and Discussions}
\subsubsection*{Importance of skip connections}
To assess the importance of skip connections within the SynNet architecture, we defined a comparative method (CM-2) with no skip connections and compared the quality of prediction results against our proposed SynNet. The performances are reported from all the 2-defined metrics in Table \ref{table:Comparative Methods}. Overall, we observe that SynNet outperforms CM-2 in reconstruction of the respective target images. This observed improvement is owed to the introduction of skip connections for feature map concatenation from encoder to decoder which provides a zero-resistance path for gradients to flow easily from deeper layers of decoder to shallower layers of encoder, thus aiding in proper weight updates throughout the network as the training progresses.\cite{drozdzal2016importance} \cite{ecb}. Also, the transfer of feature maps from the encoder block to respective the decoder block provides additional contextual information aiding in better synthesis of the images.
\subsubsection*{Importance of pixel wise weighted loss function}
we proposed to use a weighting scheme to compute the loss function, which selectively provides high importance to the edges. To justify the use of such a weighting scheme, we set up the \ref{table:BL-2} we observe a fall of 0.06 SSIM measure for better reconstruction of the predicted images especially in the areas corresponding to edges and non-edges as shown in the \ref{fig:T2Reconstruction}.
\subsubsection*{Effect of joint loss function}
We observe that the joint action of both the cost function provides better quality of reconstruction in comparison to each acting alone. A closer look at \ref{table:Proposed loss function} for SSIM measure reveals that quality of the reconstruction can be better with the introduction of a perceptually motivated loss function along with the traditional L2 loss. The effect is much more dominant with the joint action of weighted L2 loss along with SSIM loss.
\begin{figure}[ht]
\begin{minipage}[hb]{1.0\linewidth}
  \centerline{\includegraphics[width=12.0cm]{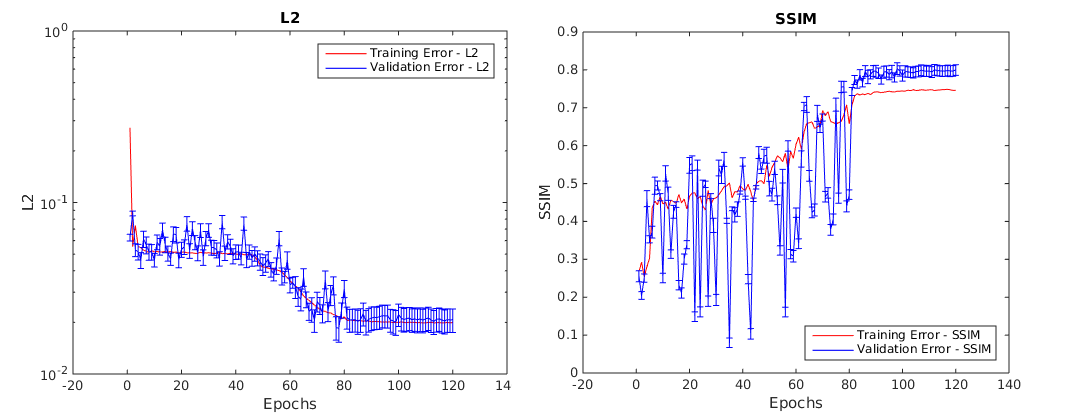}}
  \caption{\label{fig:Convergence}Convergence behavior of $L_{2} $ and SSIM loss during training}
\end{minipage}
\end{figure}

\section{Conclusion}
In this thesis, we proposed SynNet, an end to end fully convolutional neural network (F-CNN) based framework for image synthesis. We trained and validated the framework on a publicly available benchmark of expert annotated dataset.To the best of our knowledge, this is the first deep learning based fully convolutional Network architecture for image synthesis.We evaluated the novelity of the architecture, by training the SynNet for various input output configurations.We trained of SynNet using perceptual loss along with weighted L2 loss and demonstrated it to be highly effective to acquire visually pleasing results for cross modal image synthesis.We showed the skip connections are a vital element within the SynNet architecture as they provide rich contextual information and a path for easy flow of gradients across the network.The usage of unpooling layers is a very important design choice for SynNet as it aids in preserving the spatial information which is particularly critical for fine- grained image synthesis.

The proposed SynNet framework has been compared and validated against 3 state of the art deep learning methods [15],[30],[31]. Additionally, comparisons have been reported against proposed joint loss function, by validating against the traditional $L_{2}$ loss function used for image synthesis in addition to the weighted $L_{2}$ loss function. The evaluation was performed based on two standard metrics including PSNR and SSIM. We demonstrate conclusively that SynNet exhibits superior performance in these comparisons and affirm that it can be used for image synthesis task.Also we showed the novelty of the proposed SynNet architecture by extending it to SISO,MISO and MIMO configurations.Towards the future work, we would extend the SynNet for applications like segmentation, dose calculations for radiation therapy planning and image super resolution.

%

\end{document}